\documentclass[journal]{IEEEtran}

\ifCLASSINFOpdf
\else
   \usepackage[dvips]{graphicx}
\fi
\usepackage{url}

\usepackage{graphicx}
\usepackage{cite}
\usepackage{epsfig}
\usepackage{amsmath}
\usepackage{amssymb}
\usepackage{comment}
\usepackage{booktabs}
\usepackage{multirow}

\usepackage{array, caption, threeparttable}
\usepackage{caption}

\renewcommand\appendix{\setcounter{secnumdepth}{-2}}

\begin{document}

\title{AutoHR: A Strong End-to-end Baseline for Remote Heart Rate Measurement with Neural Searching}

\author{Zitong Yu, Xiaobai Li, Xuesong Niu, Jingang Shi and Guoying Zhao, \IEEEmembership{Senior Member, IEEE}
\thanks{This work was supported by the Academy of Finland for project MiGA (grant 316765), ICT 2023 project (grant 328115), and Infotech Oulu. As well, the authors wish to acknowledge CSC-IT Center for Science, Finland, for computational resources. (Corresponding author: Guoying Zhao.)}
\thanks{Z. Yu, X. Li, J. Shi and G. Zhao are with the Center for Machine Vision and Signal
Analysis, University of Oulu, Oulu 90014, Finland. E-mail: \{zitong.yu, xiaobai.li,  jingang.shi, guoying.zhao\}@oulu.fi.}
\thanks{X. Niu is with the Key Laboratory of Intelligent Information Processing, Institute of Computing Technology, Chinese Academy of Sciences, Beijing 100190, China, and also with the University of Chinese Academy of Sciences, Beijing 100049, China. E-mail: xuesong.niu@vipl.ict.ac.cn.}}

\maketitle

\begin{abstract}
Remote photoplethysmography (rPPG), which aims at measuring heart activities without any contact, has great potential in many applications (e.g., remote healthcare). Existing end-to-end rPPG and heart rate (HR) measurement methods from facial videos are vulnerable to the less-constrained scenarios (e.g., with head movement and bad illumination). In this letter, we explore the reason why existing end-to-end networks perform poorly in challenging conditions and establish a strong end-to-end baseline (AutoHR) for remote HR measurement with neural architecture search (NAS). The proposed method includes three parts: 1) a powerful searched backbone with novel Temporal Difference Convolution (TDC), intending to capture intrinsic rPPG-aware clues between frames; 2) a hybrid loss function considering constraints from both time and frequency domains; and 3) spatio-temporal data augmentation strategies for better representation learning. Comprehensive experiments are performed on three benchmark datasets to show our superior performance on both intra- and cross-dataset testing.

\end{abstract}

\begin{IEEEkeywords}
rPPG, remote heart rate measurement, neural architecture search, convolution.
\end{IEEEkeywords}

\IEEEpeerreviewmaketitle

\section{Introduction}

\IEEEPARstart{H}{eart} rate (HR) is an important vital sign that needs to be measured in many circumstances, especially for healthcare or medical purposes. Traditionally, the Electrocardiography (ECG) and Photoplethysmograph (PPG)~\cite{murthy2015multiple} are the two most common ways for measuring heart activities and corresponding average HR. However, both ECG and PPG sensors need to be attached to body parts, which may cause discomfort and are inconvenient for long-term monitoring. To counter for this issue, remote photoplethysmography (rPPG)~\cite{verkruysse2008remote,poh2010non,poh2010advancements,li2014remote,tulyakov2016self,de2013robust,wang2017algorithmic,park2018remote,shi2019atrial,hsu2017deep,qiu2018evm,niu2018synrhythm,niu2019rhythmnet,vspetlik2018visual,chen2018deepphys,yu2019remote1,yu2019remote2,wang2014exploiting} methods are developing fast in recent years, which target to measure heart activity remotely without any contact.

In earlier studies of remote HR measurement from facial videos, most traditional methods~\cite{verkruysse2008remote,poh2010non,poh2010advancements,li2014remote,tulyakov2016self,de2013robust,wang2017algorithmic} can be seen as a two-stage pipeline, which first extracts the rPPG signals from the detected/tracked face regions, and then estimates the corresponding average HR from frequency analysis. On one hand, most methods analyze subtle color changes on facial regions of interest (ROI). Verkruysse et al.~\cite{verkruysse2008remote} first found that rPPG could be recovered from facial skin regions using ambient light and the green channel featured the strongest rPPG clues. Poh et al.~\cite{poh2010non,poh2010advancements} utilized independent component analysis for noise removal, which is robust to the environment even using a low-cost webcam. Li et al. ~\cite{li2014remote} proposed to track the well-defined ROI for coarse rPPG signals recovery and then refine the signals via illumination rectification and non-rigid motion elimination. Tulyakov et al.~\cite{tulyakov2016self} proposed self-adaptive matrix completion for HR estimation, which captures the consistent clues among ROIs with noise reduction. On the other hand, there are few color subspace transformation methods which utilized all skin pixels for rPPG measurement, e.g., chrominance-based (CHROM)~\cite{de2013robust} and projection plane orthogonal to the skin tone (POS)~\cite{wang2017algorithmic}.

Based on the prior knowledge (e.g., ROI definition and signal processing) from traditional methods, several learning based approaches~\cite{ hsu2017deep,qiu2018evm,niu2018synrhythm,niu2019rhythmnet} are designed as non-end-to-end fashions. After extracting the rPPG signals via traditional CHROM~\cite{de2013robust}, Hsu et al.~\cite{ hsu2017deep} transformed the signals to time-frequency representation, which is cascaded with VGG15~\cite{simonyan2014very} for HR regression. Qiu et al.~\cite{qiu2018evm} extracted the features via spatial decomposition and temporal filtering in particular face ROI, and then CNN was utilized for mapping the hand-crafted features to HR value. Niu et al.~\cite{niu2018synrhythm,niu2019rhythmnet} generated the spatio-temporal map representation via aggregating the information within multiple small face ROIs, which is cascaded with ResNet18~\cite{he2016deep} for HR prediction. However, these methods need the strict preprocessing procedure and neglect the global clues outside the pre-defined ROI.

\begin{figure}
\centering
\includegraphics[scale=0.52]{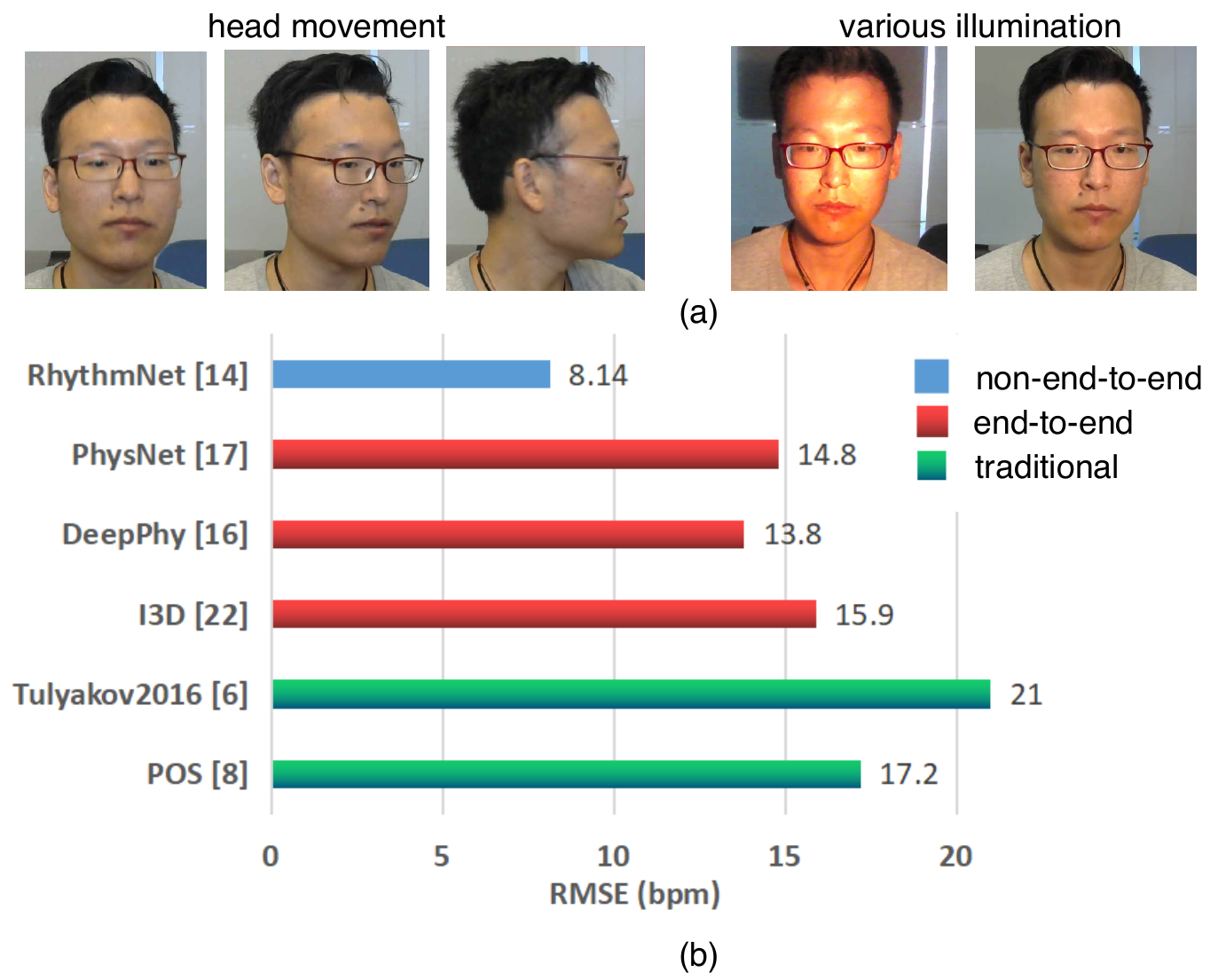}
\vspace{-0.5em}
\caption{\small{Examples and results on VIPL-HR~\cite{niu2019rhythmnet} dataset. (a) The complex senarios. (b) Root mean square error (RMSE) results. The smaller RMSE, the better performance. }}
\label{fig:Figure1}
\vspace{-0.5em}
\end{figure}

Meanwhile, a few end-to-end deep learning based rPPG methods~\cite{vspetlik2018visual,chen2018deepphys,yu2019remote1,yu2019remote2} are developed, which take the face frames as input and predict the rPPG signals or HR values directly. \v{S}petl{\'\i}k et al.~\cite{vspetlik2018visual} proposed a two-stage method (HR-CNN), which first measures the rPPG signal by a 2D CNN with frequency constraints, and then regresses the HR value via another 1D CNN. Chen and McDuff proposed convolutional attention networks for physiological measurement (DeepPhys~\cite{chen2018deepphys}), which uses the normalized difference frames as input and predicts the rPPG signal derivative. Yu et al. designed spatio-temporal networks (PhysNet~\cite{yu2019remote1} and rPPGNet~\cite{yu2019remote2}) for rPPG signal recovery, which are supervised by negative Pearson loss in the time domain. However, pure end-to-end methods are easily influenced by the complex scenarios (e.g., with head movement and various illumination conditions. See Fig.~\ref{fig:Figure1}(a) for examples). As shown in Fig.~\ref{fig:Figure1}(b), the end-to-end deep learning based rPPG methods (I3D~\cite{carreira2017quo}, DeepPhys~\cite{chen2018deepphys} and PhysNet~\cite{yu2019remote1}) fall behind the state-of-the-art non-end-to-end method (RhythmNet~\cite{niu2019rhythmnet}) by a large margin.

In this letter, we aim to explore these two questions: \textbf{1) Why the existing end-to-end deep learning based rPPG methods perform poorly in less-constrained scenarios? 2) Can end-to-end networks generalize well even in less-constrained scenarios?} Our contribution includes:

\begin{itemize}
\setlength\itemsep{-0.1em}
    \item We find out three key factors (i.e., network architecture, loss function and data augmentation strategy) influencing the robustness and generalization ability of end-to-end rPPG networks heavily.
    
    \item We propose the AutoHR, which consists of a powerful searched backbone with the Temporal Difference Convolution (TDC). To our best knowledge, it is the first time to discover well-suited backbone via Neural Architecture Search (NAS) for remote HR measurement. 

    \item We propose a hybrid loss function with both time and frequency constrains, forcing the AutoHR to learn rPPG-aware features. Besides, we present spatio-temporal data augmentation strategies for better representation learning.

    \item We conduct intra- and cross-dataset tests  and show that the AutoHR achieves superior or on par state-of-the-art performance, which can be treated as a strong end-to-end baseline for rPPG research community.
    
\end{itemize}

\section{Methodology}

\subsection{Temporal Difference Convolution}
\label{TDC}
 
\begin{figure}
\centering
\includegraphics[scale=0.21]{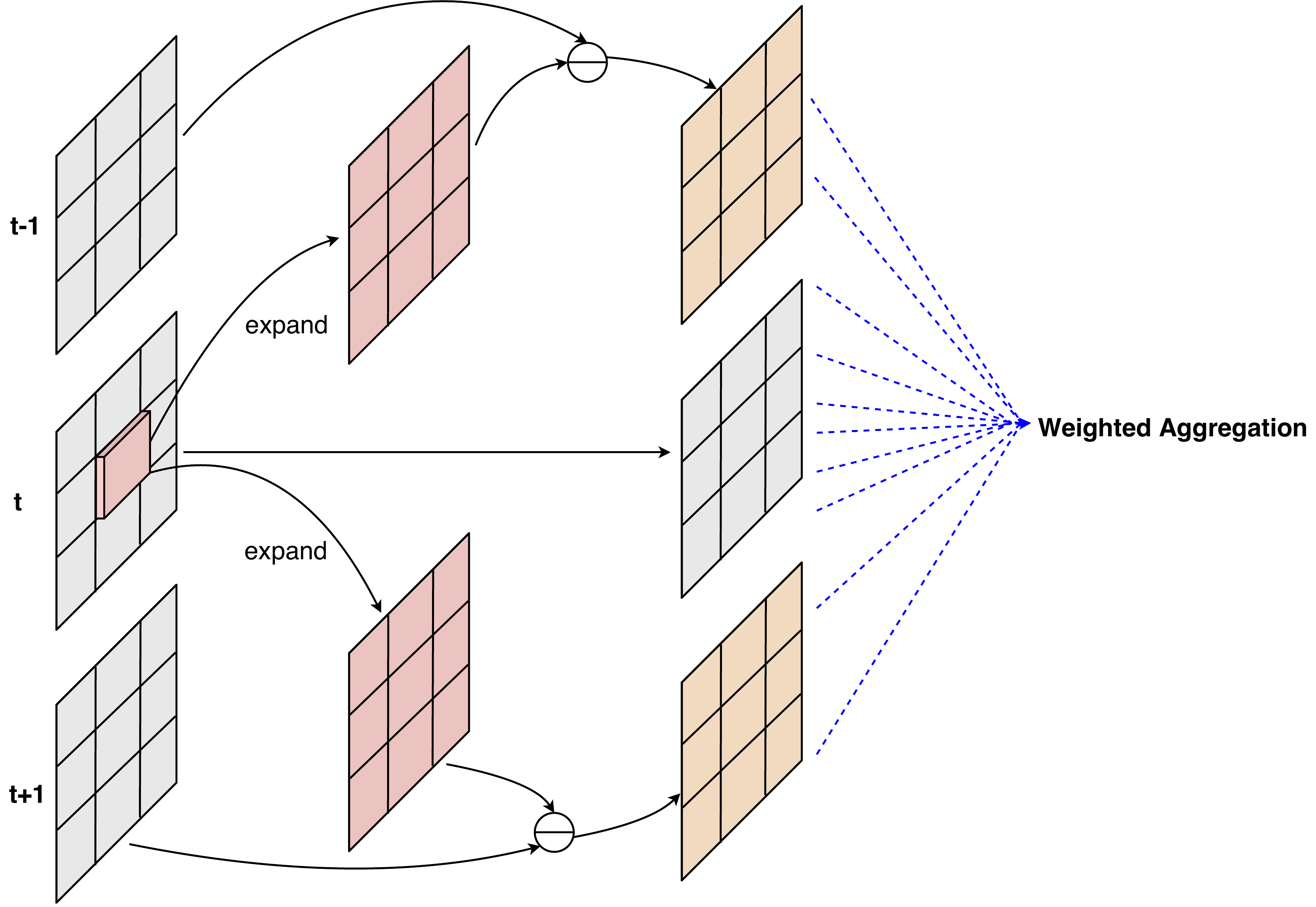}
\vspace{-0.1em}
\caption{\small{Temporal difference convolution. }}
\label{fig:TDC}
\vspace{-2.5em}
\end{figure}

Temporally normalized frame difference~\cite{wang2014exploiting} is proved to be robust for rPPG recovery in motion and bad illumination scenarios. In DeepPhys~\cite{chen2018deepphys}, a new type of normalized frame difference is designed as the network input. However, some important information will be lost due to the normalized clipping, which limits the capability of representation learning. Here we adopt the original RGB frames as input and design a novel Temporal Difference Convolution (TDC) for describing the temporal differences in feature levels. For simplicity, we describe TDC with 3$\times$3$\times$3 kernel and channel number 1. The case with larger kernel size and channel number is analogous. $w(i)$ denotes the learnable weight at local position $i$. As illustrated in Fig.~\ref{fig:TDC}, besides weighting the spatial information in the local region $\mathcal{R}_{t}$ at current time $t$, the TDC also aggregates the temporal difference clues within local temporal regions $\mathcal{R}_{t-1}$ and $\mathcal{R}_{t+1}$, which can be formulated as

\begin{equation} \small
\begin{split}
O(p^{t}_0)
=  \sum_{p^{t-1}_n\in \mathcal{R}_{t-1}}w(p^{t-1}_n)\cdot (I(p^{t-1}_0+p^{t-1}_n)-\theta \cdot I(p^{t}_0))&\\
+ \sum_{p^{t+1}_n\in \mathcal{R}_{t+1}}w(p^{t+1}_n)\cdot (I(p^{t+1}_0+p^{t+1}_n)-\theta \cdot I(p^{t}_0))&\\
+  \sum_{p^{t}_n\in \mathcal{R}_{t}}w(p^{t}_n)\cdot I(p^{t}_0+p^{t}_n),& \\
\end{split}
\label{eq:TDC}
\end{equation}
where $p^{t}_0$ denotes current location at time $t$ on both input $I$ and output $O$ feature maps while $p^{t-1}_n$, $p^{t}_n$ and $p^{t+1}_n$ enumerate the locations in $\mathcal{R}_{t-1}$, $\mathcal{R}_{t}$ and $\mathcal{R}_{t+1}$, respectively. For instance, local receptive field region $\mathcal{R}_{t-1}$, $\mathcal{R}_{t}$ and $\mathcal{R}_{t+1}$ can be represented as $\left \{  (-1,-1),(-1,0),\cdots,(0,1),(1,1)  \right \}$. The hyperparameter $\theta \in [0,1]$ tradeoffs the contribution of temporal difference. The higher value of $\theta$ means the more importance of temporal difference information. Specially, TDC degrades to vanilla 3D convolution when $\theta =0$. 

Overall, the advantages of introducing temporal difference clues into vanilla 3D convolution are in two folds: 1) cascaded with deep normalization operators, TDC is able to mimic the temporally normalized frame difference~\cite{wang2014exploiting} in feature levels, and 2) temporal central difference information provides fine-grained temporal context, which might be helpful to track the local ROIs for robust rPPG recovery.

\subsection{Backbone Search for Remote HR Measurement}
\label{NAS}

All existing end-to-end rPPG networks~\cite{vspetlik2018visual,chen2018deepphys,yu2019remote1,yu2019remote2} are designed manually, which might be sub-optimal for feature representation. In this paper, we first introduce NAS to automatically discover the best-suited backbone for the task of remote HR measurement. Our search algorithm is based on two gradient-based NAS methods \cite{liu2018darts,xu2019pc}, and more technical details can be referred to the original papers.

\begin{figure}
\centering
\includegraphics[scale=0.31]{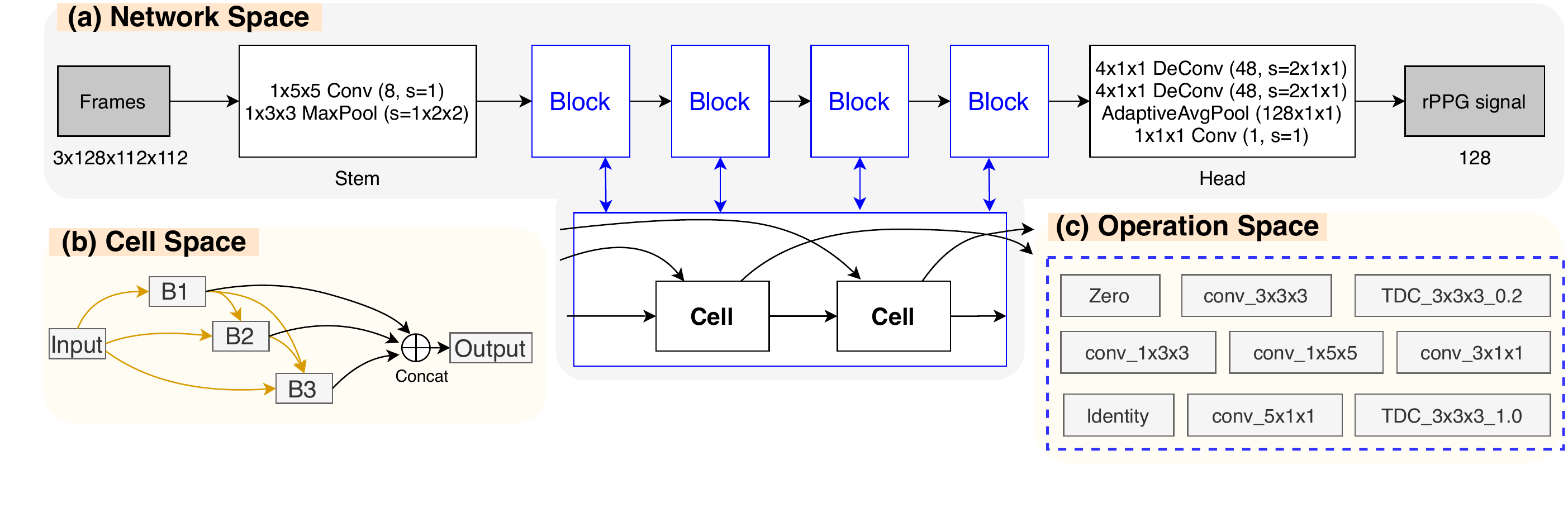}
\vspace{-1.8em}
  \caption{\small{
  Architecture search space. (a) Network space consists of one stem block, one head block, and four stacked blocks with two cells inside. There are three MaxPool layers cascaded after the first three `Block' with stride 1x2x2, 2 and 2, respectively. (b) Cell space contains 5 nodes, including an input node, three intermediate nodes B1, B2, B3 and an output node. The edge between two nodes (except output node) denotes a possible operation. (c) The operation space consists of nine candidates, where `TDC\_3x3x3\_$\theta$' means temporal difference convolution with kernel size 3 and hyperparameter $\theta$. }
  }
 
\label{fig:searchspace}
\vspace{-0.3em}
\end{figure}

As illustrated in Fig.~\ref{fig:searchspace}(a), our goal is to search for cells to form a network backbone for the rPPG recovery task. As for the cell-level structure, Fig.~\ref{fig:searchspace}(b) shows that each cell is represented as a directed acyclic graph (DAG) of $N$ nodes $\left \{ x \right \}^{N-1}_{i=0}$, where each node represents a network layer. We denote the operation space as $\mathcal{O}$, and Fig.~\ref{fig:searchspace}(c) shows nine designed candidate operations. Each edge $(i,j)$ of DAG represents the information flow from node $x_{i}$ to node $x_{j}$, which consists of the candidate operations weighted by the architecture parameter $\alpha^{(i,j)}$. Specially, each edge $(i,j)$ can be formulated by a function $\tilde{o}^{(i,j)}$ where $\tilde{o}^{(i,j)}(x_i)=\sum_{o\in \mathcal{O}}\eta_{o}^{(i,j)}\cdot o(x_{i})$. Softmax function is utilized to relax architecture parameter $\alpha^{(i,j)}$ into operation weight $o\in \mathcal{O}$, that is $\eta_{o}^{(i,j)}=\frac{exp(\alpha_{o}^{(i,j)})}{\sum_{{o}'\in \mathcal{O}}exp(\alpha_{{o}'}^{(i,j)})}$. The intermediate node can be denoted as $x_{j}=\sum_{i<j}{\tilde{o}}^{(i,j)}(x_{i})$ and the output node $x_{N-1}$ is depth-wise concatenation of all the intermediate nodes excluding the input nodes.

In general, the architecture of the cells within four blocks are shared for robust searching. We also consider the flexible and complex setting when all the four blocks to be searched are varied. We name these two configurations as `Shared' and `Varied', respectively. In addition, we also compare the operation space with and without TDC in Section \ref{sec:ablation}.

\vspace{0.2em}

In the searching stage, $\mathcal{L}_{train}$ and $\mathcal{L}_{val}$ are denoted as
the training and validation loss respectively, which are all based on the overall loss $\mathcal{L}_{overall}$ described in Section \ref{sec:loss}. Network parameters $\Phi$ and architecture parameters $\alpha$ are learned with the following bi-level optimization problem:

\vspace{-0.5em}

\begin{equation} 
\begin{split}
&\underset{\alpha}{min} \quad \mathcal{L}_{val}(\Phi ^{*}(\alpha),\alpha ), \\
&s.t. \quad \Phi  ^{*}(\alpha)=arg  \,
\underset{\Phi }{min}   \;
\mathcal{L}_{train}(\Phi ,\alpha)
\end{split}
\label{eq:optimize}
\vspace{-2.1em}
\end{equation}

After convergence, the final discrete architecture is derived by: 1) setting $o^{(i,j)}=arg\,max_{o\in \mathcal{O},o\neq none}\,p_{o}^{(i,j)}$, and 2) for each intermediate node, choosing two incoming edges with the two largest values of $max_{o\in \mathcal{O},o\neq none}\,p_{o}^{(i,j)}$.

\subsection{Supervision in the Time and Frequency Domain}
\label{sec:loss}

Besides designing the network architecture, we also need an appropriate loss function to guide the networks. However, existing negative Pearson (NegPearson)~\cite{yu2019remote1,yu2019remote2} and signal-to-noise ratio (SNR)~\cite{vspetlik2018visual} losses only constrain in the time or frequency domains, respectively. It might be helpful for the network to learn more intrinsic rPPG features with both strong spectral distribution supervision in the frequency domain, and fine-grained signal rhythm guidance in the time domain. The loss in the time domain can be formulated as

\vspace{-0.6em}
\begin{equation} \small
\mathcal{L}_{time}=1-\frac{T\sum_{1}^{T}XY-\sum_{1}^{T}X\sum_{1}^{T}Y}{\sqrt{(T\sum_{1}^{T}X^2-(\sum_{1}^{T}X)^2)(T\sum_{1}^{T}Y^2-(\sum_{1}^{T}Y)^2)}},
\vspace{-0.3em}
\end{equation}
where \(X\) and \(Y\) indicate the predicted rPPG and ground truth PPG signals, respectively. \(T\) is the length of the signals. Inspired by SNR loss, we also treat HR estimation as a classification task via frequency transformation, which is formulated as  
\vspace{-0.1em}
\begin{equation} \small
\mathcal{L}_{fre}=CE(PSD(X),HR_{gt}),
\vspace{-0.1em}
\end{equation}
where $PSD(X)$ is the power spectral density of the predicted rPPG signal $X$, while $HR_{gt}$ denotes the ground truth HR value. $CE$ means the classical cross-entropy loss. Finally, the overall loss in the time and frequency domain is $\mathcal{L}_{overall}=\lambda\cdot\mathcal{L}_{time}+\mathcal{L}_{fre}$, where $\lambda$ is a
balancing parameter.

\subsection{Spatio-Temporal Data Augmentation}
\label{sec:DA}

Two problems hindering the remote HR measurement task caught our attention, and we propose two data augmentation strategies accordingly. On one hand, head movements could cause ROI occlusion. We propose data augmentation strategy `DA1', which is to randomly erase or cutout partial spatio-temporal tubes within a random time clip (less than 20\% spatial size and 20\% temporal length), which mimics the situation of ROI occlusion. On the other hand, HR distribution is severely unbalanced as a reversed-V shape. We propose data augmentation strategy `DA2', which is to temporally upsample and downsample the videos to generate extra training samples with extreme small or large HR values. To be specific, the videos with HR values larger than 90 bpm would be temporally interpolated twice while those with HR smaller than 70 bpm are downsampled with sampling rate 2, to simulate half and doubled heart rate, respectively.



\begin{figure}
\centering
\includegraphics[scale=0.62]{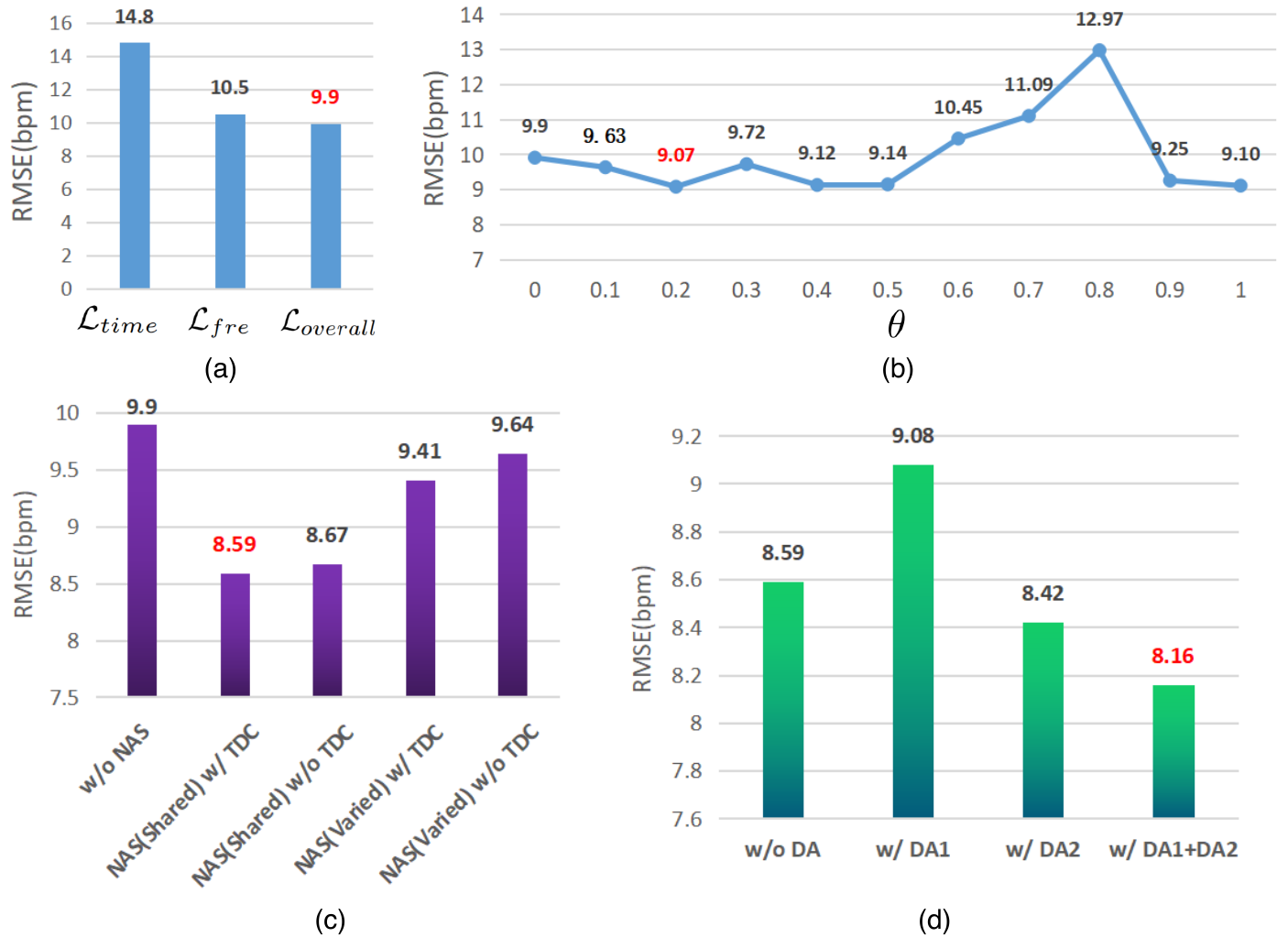}
\vspace{-1.6em}
\caption{\small{Ablation study of (a) loss functions; (b) hypterparameter $\theta$ in TDC; (c) NAS configuration; and (d) data augmentation strategies. }}
\label{fig:Ablation}
\vspace{-0.7em}
\end{figure}

\section{Experiments}
\label{sec:guidelines}

\subsection{Datasets and Metrics}
Three public datasets are employed in our experiments. The VIPL-HR~\cite{niu2019rhythmnet} and MAHNOB-HCI~\cite{soleymani2011multimodal} datasets are utilized for intra-dataset testing with subject-independent 5-fold cross-validation while the MMSE-HR~\cite{tulyakov2016self} is used for cross-dataset testing. Performance metrics for evaluating the average HR include the standard deviation (SD), the mean absolute error (MAE), the root mean square error (RMSE), and the Pearson's correlation coefficient ($r$).

\subsection{Ablation Study}
\label{sec:ablation}
All ablation studies are conducted on Fold-1 of the VIPL-HR~\cite{niu2019rhythmnet} dataset. PhysNet~\cite{yu2019remote1} is adopted as the backbone for exploring the impacts of loss functions and TDC. The NAS searched backbone is used for exploring the effectiveness of data augmentation strategies.

\textbf{Impact of Loss Functions.}\quad   Fig.~\ref{fig:Ablation}(a) shows that $\mathcal{L}_{fre}$ plays a vital role for accurate HR prediction, which reduces 4.3 bpm RMSE compared with $\mathcal{L}_{time}$. The lowest RMSE (9.9 bpm) could be obtained when supervised with  $\mathcal{L}_{overall}$, indicating the elaborate supervision in both time and frequency domains enables the model to learn robust rPPG features. 

\textbf{Impact of $\theta$ in TDC.}\quad 
According to Eq.~(\ref{eq:TDC}), $\theta$ controls the contribution of the temporal difference information. As illustrated in Fig.~\ref{fig:Ablation}(b), compared with the vanilla 3D convolution (i.e., $\theta$=0), TDC achieves lower RMSE in most cases, indicating the temporal difference clues are helpful for HR measurement. Specially, we consider $\theta=0.2$ and $\theta=1$ in our search operation space because of their better performance (RMSE=9.07 and 9.1 bpm, respectively).

\textbf{Impact of NAS Configuration.}\quad 
As shown in Fig.~\ref{fig:Ablation}(c), the searched backbones always perform better than that designed manually without NAS. It is interesting that NAS with shared cells is more likely to search excellent architectures than that with varied cells, which might be caused by the inefficient search algorithm and limited amounts of data. Moreover, better-suited networks could be searched with TDC operators.

\textbf{Impact of Data Augmentation.}\quad 
Fig.~\ref{fig:Ablation}(d) illustrates the evaluation results of various data augmentation strategies. It is surprised that only with `DA1' (without `DA2'), it harms the performance slightly (0.5 bpm RMSE increased). The reason might be that the model learns rPPG-unrelated features due to the random spatio-temporal cutouts especially under the conditions of unbalanced data distribution. However, with the help of `DA2' to enrich the samples with extreme HR values, the strategy `DA1+DA2' improves the performance by 5\%.

\begin{table}[t]\small
\vspace{-0.3em}
\centering
\caption{Intra-dataset results on VIPL-HR. The symbols $\blacktriangle$, $\blacklozenge$ and $\star$ denote traditional, non-end-to-end learning based and end-to-end learning based methods, respectively. Best results are marked in \textbf{bold} and second best in \underline{underline}.} \label{tab:ResultsVIPL}
\resizebox{0.45\textwidth}{!} {\begin{tabular}{l c c c c} 
 \toprule
 Method & SD (bpm) & MAE (bpm)  & RMSE (bpm)  & $r$\\
 \midrule
 Tulyakov2016~\cite{tulyakov2016self}$\blacktriangle$ & 18.0 & 15.9 & 21.0 &  0.11\\
 POS~\cite{wang2017algorithmic}$\blacktriangle$ & 15.3 & 11.5 & 17.2 & 0.30 \\
 CHROM~\cite{de2013robust}$\blacktriangle$ & 15.1 & 11.4 & 16.9 & 0.28 \\ 
 \midrule
 RhythmNet~\cite{niu2019rhythmnet}$\blacklozenge$ & \textbf{8.11} & \textbf{5.30} & \textbf{8.14} &  \textbf{0.76}\\
 \midrule
 I3D~\cite{carreira2017quo}$\star$ & 15.9  & 12.0 & 15.9 & 0.07\\
  PhysNet~\cite{yu2019remote1}$\star$ & 14.9 & 10.8 & 14.8 & 0.20 \\
 DeepPhys~\cite{chen2018deepphys}$\star$ & 13.6 & 11.0  & 13.8 & 0.11 \\ 

 \textbf{AutoHR (Ours)$\star$} & \underline{8.48
} & \underline{5.68} & \underline{8.68} & \underline{0.72}\\
 \bottomrule
 \end{tabular}}
\vspace{-1.5em}
\end{table}

\vspace{-1.3em}
\subsection{Intra-dataset Testing}

\textbf{Results on VIPL-HR.} \quad   As shown in Table~\ref{tab:ResultsVIPL}, all three traditional methods (Tulyakov2016~\cite{tulyakov2016self}, POS~\cite{wang2017algorithmic} and CHROM~\cite{de2013robust}) perform poorly because of the complex scenarios (e.g., large head movement and various illumination) in the VIPL-HR dataset. Similarly, the existing end-to-end learning based methods (e.g., PhysNet~\cite{yu2019remote1} and DeepPhys~\cite{chen2018deepphys}) predict unreliable HR values as their Pearson's correlation coefficient are quite low ($r\le0.2$). In contrast, our proposed AutoHR achieves comparable performance with state-of-the-art non-end-to-end learning based method RhythmNet~\cite{niu2019rhythmnet}, which can be regarded as a strong end-to-end baseline for the challenging VIPL-HR dataset. Note that RhythmNet needs the strict and heavy preprocessing procedure to eliminate external disturbances while our AutoHR learns the intrinsic rPPG-aware features automatically without any preprocessing.

\textbf{Results on MAHNOB-HCI.} \quad   
We also evaluate our method on the MAHNOB-HCI dataset, which is widely used in HR measurement. The video samples are challenging because of the high compression rate and spontaneous motions, caused by facial expressions for example. As shown in Table~\ref{tab:ResultsMAHNOB}, the proposed AutoHR achieves the lowest MAE (3.78 bpm) among the traditional and end-to-end learning methods, which indicates the robustness of the learned rPPG features. Our performance is on par with the latest non-end-to-end learning based method RhythmNet~\cite{niu2019rhythmnet}. It implies that with excellent architecture and sufficient supervision, the end-to-end learning fashion is possible for robust HR measurement.

\begin{table}[t]\small
\vspace{-0.3em}
\centering
\caption{Intra-dataset results on MAHNOB-HCI.} \label{tab:ResultsMAHNOB}
\resizebox{0.45\textwidth}{!} {\begin{tabular}{l c c c c} 
 \toprule
 Method & SD (bpm) & MAE (bpm)  & RMSE (bpm)  & $r$\\
 \midrule
 Poh2011~\cite{poh2010advancements}$\blacktriangle$ & 13.5 & - & 13.6 & 0.36 \\ 
 CHROM~\cite{de2013robust}$\blacktriangle$ & - & 13.49 & 22.36 & 0.21 \\
 Li2014~\cite{li2014remote}$\blacktriangle$ & 6.88 & - & 7.62 & 0.81\\
 Tulyakov2016~\cite{tulyakov2016self}$\blacktriangle$ & 5.81 & 4.96 & 6.23 & 0.83\\
  \midrule
 SynRhythm~\cite{niu2018synrhythm}$\blacklozenge$ & 10.88 & - & 11.08 & - \\ 
 RhythmNet~\cite{niu2019rhythmnet}$\blacklozenge$ & \textbf{3.99} & - & \textbf{3.99}  & \textbf{0.87} \\ 
 \midrule
 HR-CNN~\cite{vspetlik2018visual}$\star$ & - & 7.25 & 9.24 & 0.51 \\
  rPPGNet~\cite{yu2019remote2}$\star$ & 7.82 & 5.51 & 7.82 & 0.78 \\
  DeepPhys~\cite{chen2018deepphys}$\star$ & - & \underline{4.57} & - & -\\
 \textbf{AutoHR (Ours)$\star$} & \underline{4.73} & \textbf{3.78} & \underline{5.10} & \underline{0.86}\\
 \bottomrule
 \end{tabular}}
\end{table}

\begin{table}[t]\small
\vspace{-0.3em}
\centering
\caption{Cross-dataset results on MMSE-HR.} \label{tab:ResultsMMSE}
\resizebox{0.35\textwidth}{!} {\begin{tabular}{l c c c c} 
 \toprule
 Method & SD (bpm)  & RMSE (bpm)  & $r$\\
 \midrule
 Li2014~\cite{li2014remote}$\blacktriangle$ & 20.02  & 19.95 & 0.38\\
 CHROM~\cite{de2013robust}$\blacktriangle$ & 14.08 & 13.97 & 0.55 \\
 Tulyakov2016~\cite{tulyakov2016self}$\blacktriangle$ & 12.24 & 11.37 & 0.71\\
 \midrule
 RhythmNet~\cite{niu2019rhythmnet}$\blacklozenge$ & \underline{6.98} & \underline{7.33} & \underline{0.78} \\ 
 \midrule
 PhysNet~\cite{yu2019remote1}$\star$ & 12.76 & 13.25 & 0.44 \\
 \textbf{AutoHR (Ours)$\star$} & \textbf{5.71}  & \textbf{5.87} & \textbf{0.89}\\

 \bottomrule
 \end{tabular}}
 \vspace{-0.5em}
\end{table}

\vspace{-0.5em}
\subsection{Cross-dataset Testing}
 In this experiment, the VIPL-HR database is used for training and all videos in the MMSE-HR database are directly used for testing. It is clear that the proposed AutoHR also generalizes well in unseen domain. It is worth noting that AutoHR achieves the highest $r$ (0.89) among the traditional, non-end-to-end and end-to-end learning based methods, which signifies 1) our predicted HRs are highly correlated with the ground truth HRs, and 2) our model learns domain-invariant intrinsic rPPG-aware features.

\vspace{-0.5em}
\section{Conclusion}
\vspace{-0.2em}

In this letter, we explore three main factors (i.e., network architecture, loss function and data augmentation) influencing the performance of the 3DCNN based end-to-end framework (e.g., PhysNet~\cite{yu2019remote1}) for remote HR measurement. The proposed AutoHR generalizes well even in less-constrained scenarios, which is promising to be a strong end-to-end baseline for rPPG research community.

\bibliographystyle{IEEEtran}
\bibliography{IEEEabrv,reference}

\clearpage

\thispagestyle{empty}
\appendix

\section{Appendix A: Dataset Details}

The VIPL-HR~\cite{niu2019rhythmnet} dataset has 3130 videos recorded from 107 subjects. The ground truth  HR and the PPG signals are extracted from the finger BVP sensors. The MAHNOB-HCI~\cite{soleymani2011multimodal} dataset is one of the most widely used benchmark for remote HR evaluations. It includes 527 videos from 27 subjects. We use the EXG2 signals as the ground truth ECG signal in evaluation. We follow the same routine as the previous work~\cite{li2014remote} and use 30 seconds clip (frames 306 to 2135) of each video.
The MMSE-HR~\cite{tulyakov2016self} dataset consists of 102 videos from 40 subjects, and the ground truth HR values are provided inside, which are computed from ECG signals.

\section{Appendix B: Implementation Details}

 Our proposed method is implemented with Pytorch. For each video clip, we use the MTCNN face detector ~\cite{zhang2016joint} to crop the enlarged face area at the first frame and fix the region through the following frames. The setting $\lambda=0.2$ is utilized for loss tradeoff. 

\textbf{Training and Testing Setting.}\quad 
 In the training stage, we randomly sample face clips with size 3$\times$160$\times$112$\times$112 (Channel$\times$Time$\times$Height$\times$Width) as the network inputs. The models are trained with Adam optimizer and the initial learning rate (lr) and weight decay (wd) are 1e-4 and 5e-5, respectively. We train models with maximum 15 epochs. The batch size is 4 on two P100 GPUs. In the testing stage, similar to ~\cite{niu2019rhythmnet}, we uniformly separate 30-second videos into three short clips with 10 seconds, and then the video-level HR is calculated via averaging the HR from three short clips.   

\textbf{Searching Setting.}  Similar to \cite{xu2019pc}, partial channel connection and edge normalization are adopted. In the training stage, we randomly sample face clips with size 3$\times$128$\times$112$\times$112. The initial channel number is 8, which doubles after searching. Adam optimizer with lr=1e-4 and wd=5e-5 is utilized when training the model weights. The architecture parameters are trained with Adam optimizer with lr=6e-4 and wd=1e-3. We search 12 epochs on Fold-1 of the VIPL-HR dataset with batchsize 2 while architecture parameters are not updated in the first five epochs. The whole searching process costs ten days on a P100 GPU.

\begin{figure}[h]
\centering
\includegraphics[scale=0.26]{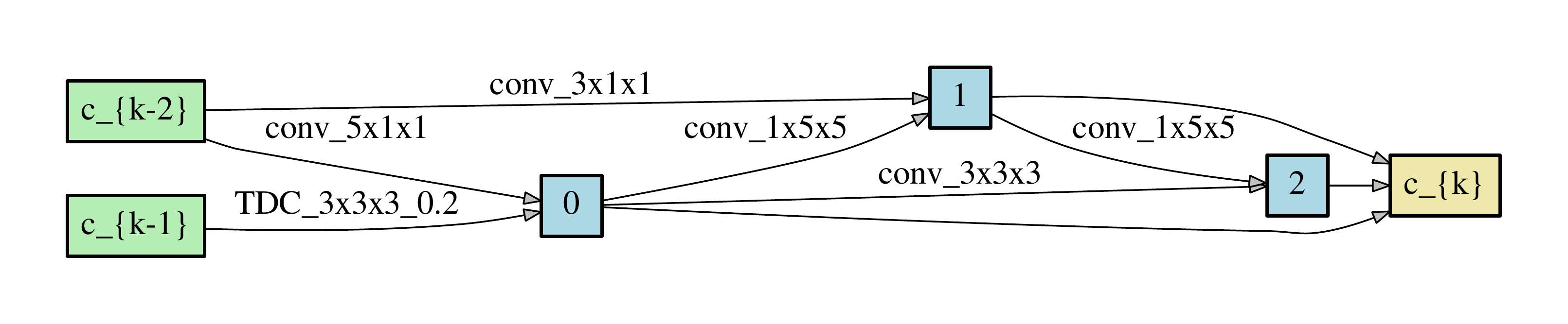}
\caption{\small{The searched cell in AutoHR. }}
\label{fig:cell}
\end{figure}

\section{Appendix C: The searched architecture}
The architecture of the searched cell of the proposed AutoHR is illustrated in Fig.~\ref{fig:cell}. It seems that the former nodes, i.e., the intermediate node 0 and node 1, prefer to utilize temporal convolutions like `conv\_3x1x1' and `conv\_5x1x1' while the latter node 2 favors to adopt spatial convolution like `conv\_1x5x5'. It might inspire the rPPG community for the further task-aware network design.

\section{Appendix D: Visualization}

\begin{figure}[h]
\centering
\includegraphics[scale=0.27]{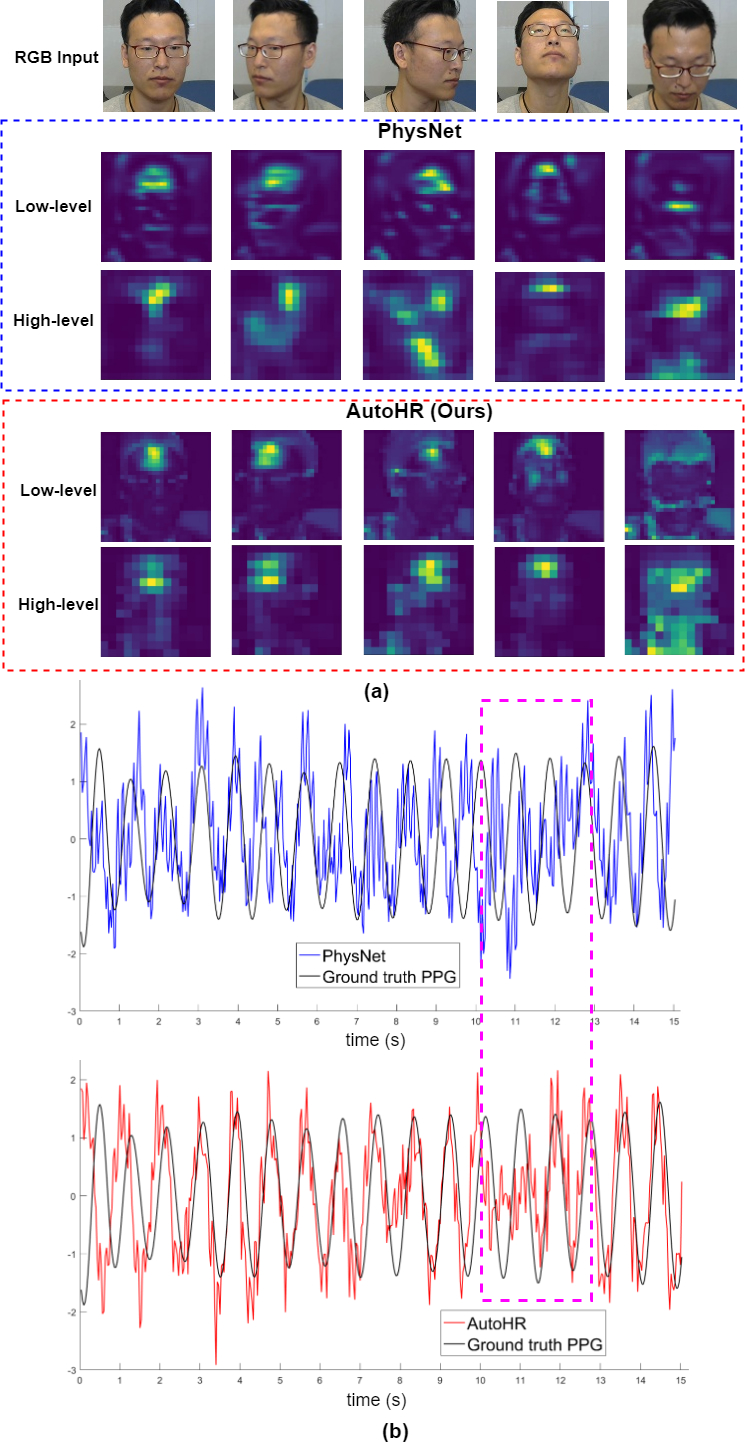}
\vspace{-0.5em}
\caption{\small{Visualization of PhysNet~\cite{yu2019remote1} and AutoHR. (a) Neural activations of the low-level and high-level features. (b) Predicted rPPG signals in the scenario with head movement. }}
\label{fig:visualization}
\end{figure}

Here we show some visualization of the feature activations and predicted rPPG signals from hard samples. As shown in Fig.~\ref{fig:visualization}(a), in terms of the low-level features (after the first `Block'), both PhysNet~\cite{yu2019remote1} and AutoHR focus more on the forehead regions, which is in accordance with the priori knowledge mentioned in~\cite{verkruysse2008remote}. However, with the head movement, the high-level features (after the third `Block') from PhysNet are not stable while those from AutoHR are still robust in tracking the particular regions. It is worth noting that features from both PhysNet and AutoHR are chaotic when the head bows (see the last column).

Fig.~\ref{fig:visualization}(b) shows the predicted rPPG signals in the scenario with head movement (slow movement at the beginning while fast movement in the later stage). It is clear that the rPPG signals predicted from AutoHR are highly correlated with ground truth PPG signals. However, due to the fast head movement (between 10 and 13 seconds along the time axes), the predicted signals are quite noisy, which indicates that the end-to-end learning based methods are vulnerable to such quick and large pose changes.

\end{document}